\documentclass[sigconf]{acmart}
\usepackage{booktabs}
\usepackage{supertabular}
\usepackage{array}
\usepackage{supertabular}
\usepackage{longtable}
\usepackage{booktabs}
\usepackage{array}
\usepackage{siunitx}
\usepackage{booktabs}
\usepackage{tabularx}
\usepackage{siunitx}
\usepackage{graphicx}

\usepackage{tabularray}
\UseTblrLibrary{booktabs,siunitx} 
\newcolumntype{C}[1]{>{\centering\arraybackslash}p{#1}}
\AtBeginDocument{%
  }

\setcopyright{acmlicensed}
\copyrightyear{2018}
\acmYear{2018}
\acmDOI{XXXXXXX.XXXXXXX}
\acmConference[Conference acronym 'XX]{Make sure to enter the correct
  conference title from your rights confirmation email}{June 03--05,
  2018}{Woodstock, NY}
\acmISBN{978-1-4503-XXXX-X/2018/06}

\begin{document}

\title{Silicon Bureaucracy and AI Test-Oriented Education: \\Contamination Sensitivity and Score Confidence in LLM Benchmarks}

\author{Yiliang Song}
\authornote{Equal contribution; work done while interning at TeleAI.}
\affiliation{%
  \institution{Institute of Artificial Intelligence (TeleAI), China Telecom}
  \country{}
}
\affiliation{%
  \institution{Guangxi Normal University}
  \country{China}
}

\author{Hongjun An}
\authornotemark[1]
\affiliation{%
  \institution{Institute of Artificial Intelligence (TeleAI), China Telecom}
  \country{}
}
\affiliation{%
  \institution{Northwestern Polytechnical University}
  \country{China}
}

\author{Jiangan Chen}
\affiliation{%
  \institution{Guangxi Normal University}
  \country{China}
}

\author{Xuanchen Yan}
\affiliation{%
  \institution{Northwestern Polytechnical University}
  \country{China}
}

\author{Huan Song}
\affiliation{%
  \institution{Institute of Artificial Intelligence (TeleAI), China Telecom}
  \country{China}
}

\author{Jiawei Shao}
\affiliation{%
  \institution{Institute of Artificial Intelligence (TeleAI), China Telecom}
  \country{China}
}

\author{Xuelong Li}
\authornote{Corresponding author.}
\email{xuelong\_li@ieee.org}
\affiliation{%
  \institution{Institute of Artificial Intelligence (TeleAI), China Telecom}
  \country{China}
}

\renewcommand{\shortauthors}{Trovato et al.}

\begin{abstract}
Public benchmarks increasingly govern how large language models (LLMs) are ranked, selected, and deployed. 
We frame this benchmark-centered regime as Silicon Bureaucracy and AI Test-Oriented Education, and argue that it rests on a fragile assumption: that benchmark scores directly reflect genuine generalization. 
In practice, however, such scores may conflate exam-oriented competence with principled capability, especially when contamination and semantic leakage are difficult to exclude from modern training pipelines. 
We therefore propose an audit framework for analyzing contamination sensitivity and score confidence in LLM benchmarks. 
Using a router-worker setup, we compare a clean-control condition with noisy conditions in which benchmark problems are systematically deleted, rewritten, and perturbed before being passed downstream. 
For a genuinely clean benchmark, noisy conditions should not consistently outperform the clean-control baseline. 
Yet across multiple models, we find widespread but heterogeneous above-baseline gains under noisy conditions, indicating that benchmark-related cues may be reassembled and can reactivate contamination-related memory. 
These results suggest that similar benchmark scores may carry substantially different levels of confidence. 
Rather than rejecting benchmarks altogether, we argue that benchmark-based evaluation should be supplemented with explicit audits of contamination sensitivity and score confidence.
\end{abstract}


\begin{teaserfigure}
  \centering
  \includegraphics[width=0.8\textwidth]{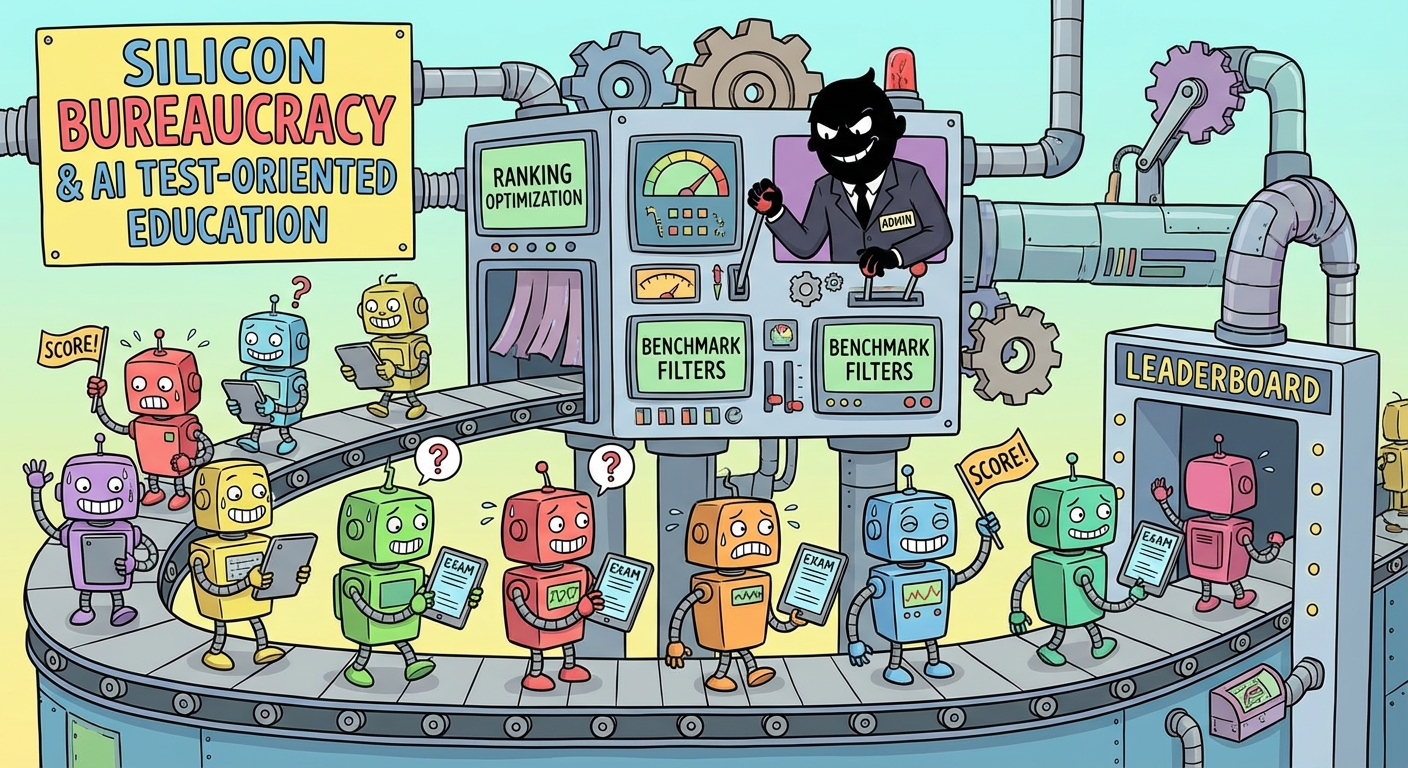}
  \caption{Silicon Bureaucracy and AI Test-Oriented Education.}
  \label{fig:teaser}
\end{teaserfigure}

\received{20 February 2007}
\received[revised]{12 March 2009}
\received[accepted]{5 June 2009}

\maketitle

\section{Introduction}
The development of large language models (LLMs) is increasingly organized around the scores, rankings, and leaderboards produced by public benchmarks \citep{kwan2024m4le,bai2024mtbench101,sun2024scieval,chen2024ragbenchmark,an2026aiflow,shao2025ai,song2025theoretical,song2026ruyi2,yuan2025information}. In academia, industry, and the broader public sphere, benchmark scores are no longer merely technical indicators for research communication \citep{alzahrani2024benchmarks,wang2026metaeval,johri2025craftmd}. They have gradually become evaluative criteria with real institutional consequences: models are examined, ranked, filtered, and treated as reference objects in procurement, investment, and governance decisions \citep{alzahrani2024benchmarks,pham2025truth,bean2025constructvalidity,wang2026metaeval}. In this sense, benchmarks have shifted from research tools to institutionalized examination and selection devices \citep{pham2025truth,bean2025constructvalidity,cais2026hle}. The ranking, certification, and comparison logic built around them reflects a pattern that deserves critical reflection, namely, Silicon Bureaucracy and AI Test-Oriented Education.

Yet this institutionalized mode of evaluation rests on a strong assumption: a high benchmark score reliably indicates stronger and more genuine generalization ability \citep{bean2025constructvalidity,huang2025thinkbench,wang2026metaeval}. This paper argues that the assumption is not robust \citep{alzahrani2024benchmarks,bean2025constructvalidity,sun2025emperorsnewclothes}. Current LLM benchmarks often measure not one pure notion of ``true generalization,'' but a mixture of two qualitatively different capacities \citep{bean2025constructvalidity,pham2025truth}. One is exam-oriented competence: under fixed answering formats, input--output conventions, and judging rules, the model can produce the correct answer \citep{pham2025truth,alzahrani2024benchmarks}. The other is the ability that remains after contamination and semantic leakage are excluded as much as possible, namely principled understanding and transfer to genuinely unseen tasks \citep{zhang2024darg,huang2025thinkbench,sun2025emperorsnewclothes}. The former is closer to ``can it answer,'' while the latter is closer to ``can it truly generalize'' \citep{bean2025constructvalidity,huang2025thinkbench}. In practice, however, many benchmarks compress both into a single score and then treat that score as a proxy for overall capability \citep{alzahrani2024benchmarks,bean2025constructvalidity,zhang2024gradeschool,liu2023codecorrect,johri2025craftmd}.

A central reason why this interpretation is unstable is that benchmark-related information is extremely difficult to remove from the training pipeline \citep{dekoninck2024constat,ni2025trainingbenchmark,sun2025emperorsnewclothes}. LLM training data are massive and heterogeneous \citep{ni2025trainingbenchmark,sun2025emperorsnewclothes}, and after repeated crawling, cleaning, distillation, synthesis, and alignment, it is hard to guarantee that benchmark questions, answers, or closely related variants never entered training \citep{dekoninck2024constat,ni2025trainingbenchmark}. Exact deduplication may still miss paraphrases, solution fragments, discussions, distilled samples, or synthetic variants that remain semantically close to the originals \citep{dekoninck2024constat,sun2025emperorsnewclothes}. Post-training blocking of original questions or keywords may also fail to prevent semantically neighboring inputs or the indirect activation created by aggregating partial clues \citep{dekoninck2024constat,ni2025trainingbenchmark}. The benchmark problem, therefore, is not only whether exact test items appeared in training, but also whether the model encountered generalized information sufficient to point toward the correct answer \citep{dekoninck2024constat,sun2025emperorsnewclothes}. Moreover, because leaderboard competition, promotion, and selection increasingly depend on benchmark performance, model development may contain latent incentives to optimize for exam success \citep{alzahrani2024benchmarks,pham2025truth}. This further blurs the boundary between optimizing for benchmark performance and optimizing for genuine generalization \citep{pham2025truth,bean2025constructvalidity}. As a result, a high score may reflect not only stronger true generalization, but also stronger exploitation of contamination-related signals \citep{dekoninck2024constat,sun2025emperorsnewclothes}.

Motivated by these concerns, this paper proposes an audit framework for interpreting benchmark scores and assessing their credibility \citep{song2026creditaudit,bouchard2026uqlm,sokol2024benchmarkcards,wang2026metaeval}. Rather than asking only which model scores higher, we ask whether performance shows an anomalous pattern when the information in a problem is systematically deleted, rewritten, and perturbed with noise. To study this, we model a single system as upstream routers and a downstream worker. Under a clean control condition, routers transmit the original problem as completely as possible; under noisy conditions, they delete, rewrite, and perturb it, and the aggregated outputs are then sent to the worker. If a benchmark is genuinely clean, performance under noisy conditions should at most approach the clean-control baseline, but should not persistently or systematically exceed it. Once a stable above-baseline phenomenon appears, a more plausible explanation is not that noise makes the model stronger, but that deleted, rewritten, and aggregated information has reassembled into cues capable of reactivating contamination-related memory traces \citep{dekoninck2024constat,sun2025emperorsnewclothes}. In this way, we transform the question of ``what benchmark scores actually measure'' into an audit problem that can be computed, compared, and used for model evaluation and selection \citep{song2026creditaudit,bouchard2026uqlm,wang2026metaeval}.

The contributions of this paper are threefold. First, we reinterpret LLM benchmarks from the perspective of institutionalized examination and selection, and propose the framework of Silicon Bureaucracy and AI Test-Oriented Education. Second, we introduce a router--worker-based audit method that identifies sensitivity to potential contamination cues by comparing deviations between clean-control and noisy conditions, especially above-baseline gains that should not systematically occur in theory. Third, across multiple models, we show that such anomalous gains are widespread but heterogeneous, implying that even similar benchmark scores may differ substantially in credibility. Accordingly, we do not claim that benchmarks are entirely invalid; rather, we argue that benchmark scores in the LLM era should be reinterpreted and supplemented with explicit credibility auditing.

\section{Related Work}

The problem addressed in this paper lies at the intersection of three lines of research: one focuses on LLM benchmarks and leaderboard-based evaluation, another on data contamination, deduplication, and benchmark leakage, and the third on model stability and selectability under different protocols and interaction conditions \citep{alzahrani2024benchmarks,dekoninck2024constat,song2026creditaudit,zhu2024promptbench}. Relative to these lines of work, our goal is not simply to identify whether a particular benchmark has been leaked, nor to revisit the familiar question of which model has higher average performance. Rather, we seek to reconsider the meaning of benchmark scores themselves: once benchmarks have evolved into institutionalized examination and selection devices, what exactly do these scores measure, and how credible are they as indicators of genuine generalization ability \citep{bean2025constructvalidity,wang2026metaeval,bouchard2026uqlm}?

\subsection{LLM Benchmarks and Leaderboard-Based Evaluation}

As competition over LLM capability intensifies, benchmarks have gradually evolved from shared measurement tools among researchers into ranking tools, publicity tools, and practical bases for deployment decisions \citep{kwan2024m4le,bai2024mtbench101,ye2025toolhop,huang2025minilongbench,zhao2025abgen,sun2024scieval,chen2024ragbenchmark,cais2026hle}. Whether a model enters the top tier of a leaderboard affects not only academic reputation, but also product comparison, user perception, investment judgment, and actual procurement decisions \citep{alzahrani2024benchmarks,wang2026metaeval,johri2025craftmd}. In this process, benchmarks are no longer merely neutral technical measuring instruments; they increasingly take on institutional functions \citep{pham2025truth,bean2025constructvalidity,cais2026hle}. Scores resemble résumés, leaderboards resemble performance review tables, and high-scoring models are more likely to obtain the status of being seen as ``advanced,'' ``reliable,'' or ``deployable'' \citep{alzahrani2024benchmarks,wang2026metaeval}. Existing research and practice have paid more attention to how benchmarks can be used to quickly compare models, but have paid less attention to whether the scores on which such institutionalized comparisons rely are themselves equally credible \citep{li2025treeeval,wang2026metaeval,sokol2024benchmarkcards,bean2025constructvalidity}. It is precisely at this point that this paper takes a further step: our concern is not whether benchmarks are useful, but how benchmark scores should be reinterpreted once benchmarks have become institutionalized examination devices \citep{pham2025truth,bean2025constructvalidity}.

\subsection{Data Contamination, Deduplication, and Benchmark Leakage}

Research on benchmark contamination has mainly discussed exact question leakage, near-duplicates, train--test overlap, and the evaluation biases that follow from them \citep{dekoninck2024constat,ni2025trainingbenchmark,sun2025emperorsnewclothes}. Related work usually emphasizes deduplication, filtering, and cleaning of training corpora as important means of preventing benchmark contamination \citep{dekoninck2024constat,ni2025trainingbenchmark}. However, this paper argues that exact deduplication does not imply the disappearance of contamination \citep{sun2025emperorsnewclothes,dekoninck2024constat}. Even if the original benchmark questions themselves are removed, paraphrased texts that are semantically close to the originals, solution fragments, discussion records, distilled samples, or synthetic data may still remain in the training pipeline in the form of generalized information, and may still indirectly point toward the correct answer at evaluation time \citep{dekoninck2024constat,ni2025trainingbenchmark,sun2025emperorsnewclothes}. Furthermore, if post-training interventions only attempt to block original questions, reference answers, or keywords, they may still fail to block rewritten inputs or to prevent the reactivation of related memory traces after multiple partial clues are aggregated together \citep{dekoninck2024constat,ni2025trainingbenchmark}. Therefore, the benchmark problem should not be understood merely as whether the exact original questions entered the training set; it should also be understood as whether the model has already encountered semantically neighboring information sufficient to point toward the correct answer \citep{dekoninck2024constat,sun2025emperorsnewclothes}. What this paper emphasizes is precisely this broader and harder-to-govern form of contamination, which extends beyond exact question leakage \citep{dekoninck2024constat,sun2025emperorsnewclothes}.

\subsection{Stability Evaluation, CreditAudit, and the Interpretation of Scores}

Beyond contamination research, recent work has also begun to recognize that model selection cannot rely on a single average score alone, but must also consider model stability across different interaction protocols, prompt templates, and task organizations \citep{song2026creditaudit,zhu2024promptbench,bai2024mtbench101,an2026pua}. Research represented by CreditAudit\citep{song2026creditaudit} argues that model evaluation in engineering settings is not only about which model has higher average ability, but also about which model remains more stable under institutionalized calling conditions. This perspective directly informs the present paper. The difference is that CreditAudit is more concerned with a model's sensitivity to protocol and scenario variation, that is, protocol sensitivity, whereas this paper further asks about the sensitivity of benchmark scores to potential contamination cues, that is, contamination sensitivity \citep{song2026creditaudit,an2026pua}. Put differently, the former is more concerned with whether a model is stable, whereas this paper is further concerned with whether its benchmark score is trustworthy \citep{song2026creditaudit,bouchard2026uqlm}. Even when two models obtain similar benchmark scores, those scores do not necessarily have the same degree of credibility \citep{song2026creditaudit,an2026pua,bouchard2026uqlm}. Accordingly, model evaluation should not stop at comparing score levels alone; it should also ask to what extent those scores may have been influenced by contamination-related cues \citep{song2026creditaudit,dekoninck2024constat,bouchard2026uqlm}.

\section{Methodology and Hypotheses}

This section formalizes benchmark scores as empirical performance under an observed distribution and studies the conditions under which such scores can be interpreted as indicators of genuine generalization. We define benchmark score, score confidence, and contamination sensitivity; introduce a router--worker mechanism; and derive the theoretical judgment that, in contamination-free settings, noisy aggregation should not systematically outperform the clean baseline.

\subsection{Conceptual Framework}

Let $\mathcal{Q}$ denote the question space, $\mathcal{A}$ the answer space, and $\mathcal{B}_n=\{(q_i,a_i)\}_{i=1}^n\subset\mathcal{Q}\times\mathcal{A}$ a benchmark sample. For model $\theta$, define
\begin{equation}
\begin{aligned}
Y_i(\theta) &= \mathbf{1}\{\hat a_\theta(q_i)=a_i\}, \\
\hat s_n(\theta) &= \frac{1}{n}\sum_{i=1}^n Y_i(\theta).
\end{aligned}
\end{equation}
and let the corresponding population score under the benchmark distribution $P_B$ be
\begin{equation}
s_B(\theta)=\int_{\mathcal{Q}\times\mathcal{A}} \mathbf{1}\{\hat a_\theta(q)=a\}\,dP_B(q,a).
\end{equation}
Let $P_0$ denote an idealized target distribution under which benchmark-related contamination has been excluded as much as possible. Then the model's contamination-reduced ability is
\begin{equation}
\begin{aligned}
s_0(\theta) &= \int_{\mathcal{Q}\times\mathcal{A}} \mathbf{1}\{\hat a_\theta(q)=a\}\,dP_0(q,a), \\
\Delta(\theta) &= s_B(\theta)-s_0(\theta).
\end{aligned}
\end{equation}
We define score confidence as the credibility of the benchmark score as an indicator of genuine generalization ability, written abstractly as
\begin{equation}
\mathrm{Conf}(\theta)=\phi(|\Delta(\theta)|),
\qquad
\phi'(\cdot)<0.
\end{equation}
Thus, a higher score does not necessarily imply a higher-confidence score.

We further define contamination sensitivity as the responsiveness of model performance to contamination-related cues. Let $\lambda\ge0$ denote cue intensity and $m(\theta,\lambda)=\mathbb{E}[Y(\theta;\lambda)]$ the expected correctness rate. Then
\begin{equation}
\mathrm{CS}(\theta)=\left.\frac{\partial m(\theta,\lambda)}{\partial \lambda}\right|_{\lambda=0^+}.
\end{equation}
A larger value indicates that the benchmark score is more likely to contain non-generalization components.

\subsection{Router--Worker Mechanism}

For question $q\in\mathcal{Q}$, let $S(q)$ denote its latent task-relevant information. Under the clean-baseline condition, a single router $R_\theta^c:\mathcal{Q}\to\mathcal{Z}$ transmits the problem as completely as possible, producing $Z^c=R_\theta^c(q)$. The worker $W_\theta:\mathcal{Z}\to\mathcal{A}$ then answers on the basis of router output only:
\begin{equation}
\begin{aligned}
\hat a_\theta^c(q) &= W_\theta(R_\theta^c(q)), \\
Y^c(q,a;\theta) &= \mathbf{1}\{\hat a_\theta^c(q)=a\}, \\
\hat s_n^c(\theta) &= \frac{1}{n}\sum_{i=1}^n Y^c(q_i,a_i;\theta).
\end{aligned}
\end{equation}

Under noisy conditions, there are $m$ parallel routers $R_{\theta,1}^n,\dots,R_{\theta,m}^n$. Each router deletes, rewrites, and perturbs the original problem, producing $Z_j^n=R_{\theta,j}^n(q)$. Their outputs are aggregated by $A_m:\mathcal{Z}^m\to\mathcal{T}$ into
\begin{equation}
T_m=A_m(Z_1^n,\dots,Z_m^n),
\end{equation}
and the worker answers only on the basis of $T_m$:
\begin{equation}
\begin{aligned}
\hat a_\theta^{n,m}(q) &= W_\theta(T_m), \\
Y^{n,m}(q,a;\theta) &= \mathbf{1}\{\hat a_\theta^{n,m}(q)=a\}, \\
\hat s_n^{\,n,m}(\theta) &= \frac{1}{n}\sum_{i=1}^n Y^{n,m}(q_i,a_i;\theta).
\end{aligned}
\end{equation}

The score deviation of the noisy condition relative to the clean baseline is
\begin{equation}
\begin{aligned}
G_m(\theta) &= \hat s_n^{\,n,m}(\theta)-\hat s_n^c(\theta), \\
G_m^+(\theta) &= \max\{G_m(\theta),0\}.
\end{aligned}
\end{equation}
Whenever $G_m(\theta)>0$, the model is said to exhibit an above-baseline anomaly under router count $m$.

\subsection{Theoretical Judgment}

Let $N(q)$ denote information unrelated to the correct answer. A noisy router output can be abstractly written as
\begin{equation}
\begin{aligned}
Z_j^n &= D_j(S(q))\cup U_j, \\
D_j(S(q)) &\subseteq S(q), \\
U_j &\subseteq N(q)\cup \tilde N_j.
\end{aligned}
\end{equation}
where $D_j(\cdot)$ is a deletion operator and $\tilde N_j$ denotes exogenous perturbation. Hence the aggregated input received by the worker is
\begin{equation}
T_m\sim A_m(D_1(S(q))\cup U_1,\dots,D_m(S(q))\cup U_m).
\end{equation}

Under contamination-free conditions, let the worker's success probability be $\pi_\theta(T)=\mathbb{P}(W_\theta(T)=a\mid q,a)$, assumed to be weakly increasing in effective information and weakly decreasing in irrelevant noise. Then noisy aggregation can improve performance only through cross-router complementarity, while also introducing extra noise. This yields
\begin{equation}
\mathbb{E}[Y^{n,m}(q,a;\theta)\mid q,a]
\le
\mathbb{E}[Y^c(q,a;\theta)\mid q,a]+\varepsilon_m(q,\theta).
\end{equation}
and thus
\begin{equation}
\begin{aligned}
\mathbb{E}[G_m(\theta)] &\le \bar\varepsilon_m(\theta), \\
\bar\varepsilon_m(\theta) &= \int \varepsilon_m(q,\theta)\,dP_B(q,a).
\end{aligned}
\end{equation}
If the clean baseline is already close to full-information transmission, $\bar\varepsilon_m(\theta)$ should be small. In that case, noisy conditions may fluctuate around the baseline, but they should not systematically exceed it.

Now let $\Xi(q)$ denote a latent set of benchmark-related contamination cues, including paraphrased variants, solution fragments, discussion texts, distilled samples, synthetic variants, or semantically neighboring expressions that remain reachable even after local post-training blocking. If the overlap between aggregated text and contamination cues is measured by $\kappa(T_m,\Xi(q))$, then the worker's success probability may be written as
\begin{equation}
\pi_\theta(T_m,\Xi(q))=\pi_\theta^0(T_m)+\psi_\theta(\kappa(T_m,\Xi(q))),
\qquad
\psi_\theta'(\cdot)\ge0.
\end{equation}
As $m$ increases, partial clues from different noisy routers may accumulate, increasing $\kappa(T_m,\Xi(q))$. Then
\begin{equation}
\mathbb{E}[Y^{n,m}(q,a;\theta)\mid q,a]>\mathbb{E}[Y^c(q,a;\theta)\mid q,a]
\end{equation}
may hold for a nontrivial subset of questions, so that persistent positive $G_m(\theta)$ is more naturally interpreted as an external signal of contamination-related memory activation than as an ordinary noise effect.

\subsection{Hypotheses}

\textbf{H1. Contamination activation hypothesis.}  
If benchmark-related semantic-neighbor contamination exists during training, or if post-training interventions only block original benchmark items locally, then multi-router noisy aggregation is more likely to activate generalized memory related to the benchmark, thereby generating above-baseline anomalies under some router settings.

\textbf{H2. Heterogeneous sensitivity hypothesis.}  
Different models differ in their sensitivity to potential contamination cues. As a result, even when their benchmark scores are similar, the magnitude of anomalous gains and the breadth of violations may differ substantially.

\textbf{H3. Directional transition hypothesis.}  
If above-baseline anomalies are driven by contamination-related memory activation rather than random fluctuation, then as the number of noisy routers increases, wrong-to-correct transitions should rise, correct-to-wrong transitions should decline correspondingly, and the former should eventually exceed the latter.

\section{Experimental Design}

This section explains how the theoretical framework is translated into a reproducible and interpretable audit procedure. The emphasis is not on engineering complexity, but on constructing a clear comparison regime: under the same questions, the same models, and the same answering constraints, we compare the clean baseline with noisy conditions and use the resulting deviations to assess how sensitive benchmark scores are to potential contamination-related cues.

\subsection{Dataset and Sample Construction}

We conduct the experiments on a public benchmark consisting of multiple-choice test questions. From the test split, we draw a fixed sample of $n=100$ questions, with the random seed set to 42. All models and all clean/noisy conditions are evaluated on exactly the same question set. This design removes additional variation caused by sampling differences across runs. In other words, the experiment follows a same-question matched comparison rather than a comparison across different question sets, so that the observed score deviations can be more directly attributed to changes in the information transmission regime.

\subsection{Models and Experimental Settings}

The audit is repeated across multiple mainstream large language models. By default, the router and the worker are instantiated with the same model, so that model heterogeneity does not enter the transmission process itself. The clean baseline is defined as the setting \texttt{forward\_full} with $r=1$, that is, a single router is used and is instructed to transmit the original problem information as completely as possible. The noisy conditions are defined as the setting \texttt{noisy\_rewrite} with $r \in \{1,2,\dots,M\}$, where $r$ denotes the number of parallel noisy routers. The role of router count is not to test whether more agents make a model stronger; rather, it serves as a control over the intensity of cue aggregation. As $r$ increases, more locally deleted, rewritten, and perturbed versions of the same problem are aggregated together, making it more likely that benchmark-related semantic-neighbor cues are reassembled in the final input to the worker.

\subsection{Prompting and Answering Constraints}

Under the clean-baseline condition, the router is instructed to preserve the original problem as fully and accurately as possible, including the question stem, options, and relevant constraints, while not directly outputting the final answer. Under noisy conditions, each router is instructed to delete part of the useful information, rewrite the problem, and inject irrelevant noise. The outputs of multiple noisy routers are then aggregated and passed to the worker. In both conditions, the worker is not allowed to access the original question directly and must answer only on the basis of router outputs. The worker is also constrained to return a single option letter as the final answer. In this way, the difference between the clean and noisy conditions is restricted to the transmission regime itself, rather than to changes in answering rules or evaluation criteria.

\subsection{Evaluation Metrics}

Let the clean-baseline accuracy be the reference performance and the noisy-condition accuracy be the comparison performance. Their difference is defined as
\[
\mathrm{gain}=\mathrm{noisy\ accuracy}-\mathrm{clean\ baseline}.
\]
Whenever $\mathrm{gain}>0$, the noisy condition outperforms the clean baseline. To isolate only the above-baseline component, we further define
\[
\mathrm{positive\ excess}=\max(\mathrm{gain},0).
\]
A noisy setting is counted as a violation whenever $\mathrm{gain}>0$. For a given model, the number of noisy settings under which violations occur defines its violation breadth, which summarizes how broadly the model's benchmark score is affected by potential contamination-related cues.

In addition to score-level metrics, we examine question-level transition directions. If a question is answered incorrectly under the clean baseline but correctly under a noisy condition, it is counted as an improve transition (wrong$\to$correct). If a question is answered correctly under the clean baseline but incorrectly under a noisy condition, it is counted as a degrade transition (correct$\to$wrong). These transition metrics help distinguish whether above-baseline anomalies are more consistent with random fluctuation or with a directional process in which noisy aggregation systematically helps recover benchmark answers.

\section{Results and Analysis}

This section evaluates the three hypotheses developed above. The empirical logic proceeds in three steps. We first examine whether noisy conditions genuinely produce above-baseline anomalies relative to the clean baseline. We then study whether such anomalies are heterogeneous across models. Finally, we move to question-level transition patterns to assess whether the observed gains are better understood as random fluctuation or as a directional process in which noisy aggregation systematically improves outcomes.

\subsection{Overall violations: do noisy conditions really exceed the baseline?}

\begin{figure*}[h]
  \centering
  \includegraphics[width=\textwidth]{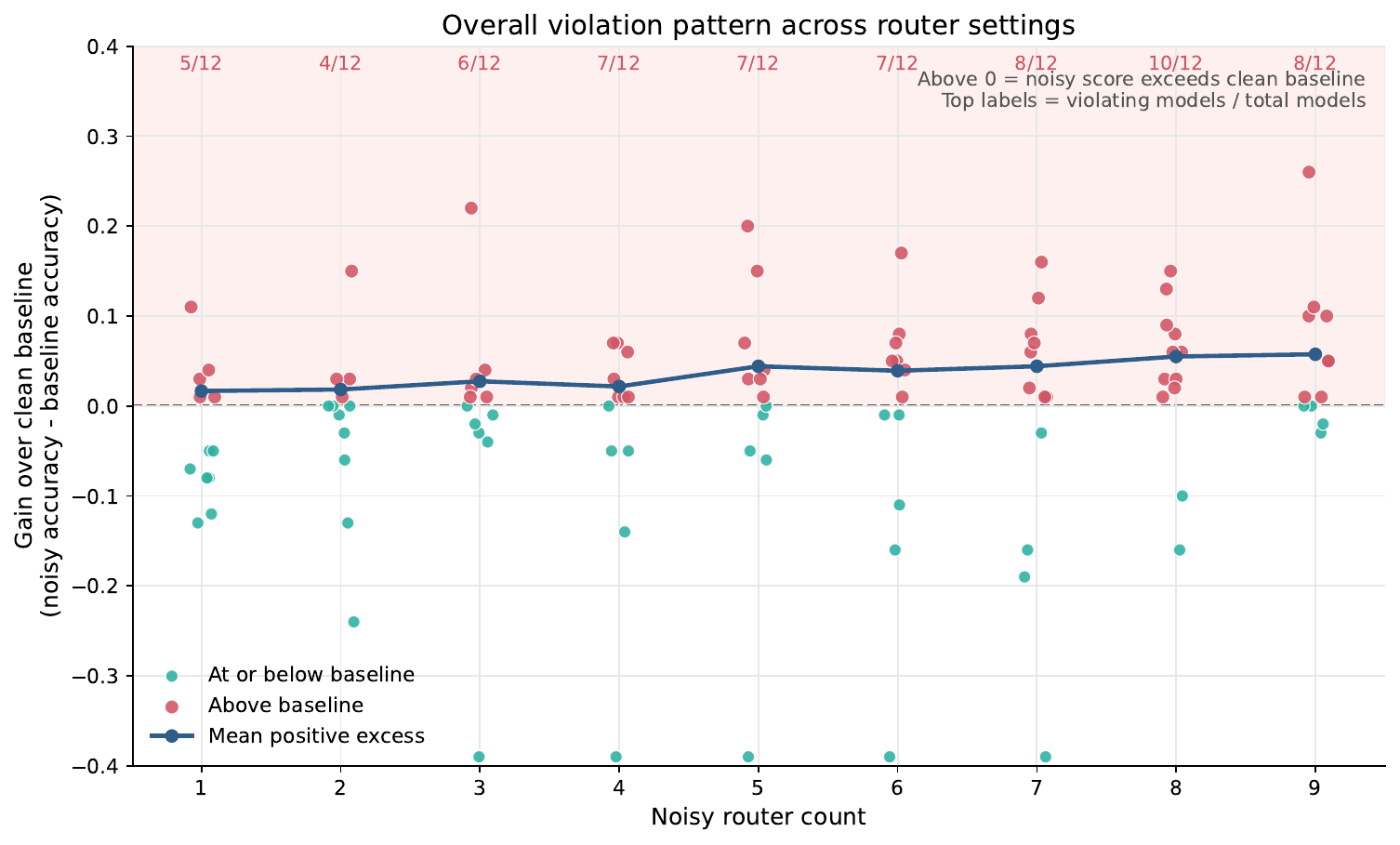}
  \caption{Overall violation pattern across router settings.}
  \label{fig:overall_violation}
\end{figure*}
Figure~\ref{fig:overall_violation} shows the overall deviation of model performance from the clean baseline under different noisy-router settings. At the aggregate level, above-baseline anomalies are not isolated outliers. As the number of noisy routers increases from 1 to 9, the number of models exceeding the clean baseline is 5/12, 4/12, 6/12, 7/12, 7/12, 7/12, 8/12, 10/12, and 8/12, respectively. The highest violation count occurs at \(r=8\), where 10 out of 12 models rise above the baseline. The mean positive excess also becomes more pronounced in higher-router regions, reaching 0.066 at \(r=8\) and 0.086 at \(r=9\), the largest value across all settings.The complete router-level summary statistics are reported in Appendix Table~\ref{tab:router_level_summary}.

These results support Hypothesis~1. If the clean baseline corresponds to the regime closest to full-information transmission, then noisy conditions should at most fluctuate around that baseline rather than persistently exceed it. The key pattern in Figure~\ref{fig:overall_violation} is therefore not that ``more routers make models stronger,'' but that higher-router settings make above-baseline anomalies both more frequent and more substantial. This is consistent with the theoretical argument that deleted, rewritten, and perturbed fragments may be recombined into semantic-neighbor cues that reactivate benchmark-related memory traces.

\subsection{Model heterogeneity: models differ in the probability and magnitude of anomalous gains}

\begin{figure*}[h]
  \centering
  \includegraphics[width=\textwidth]{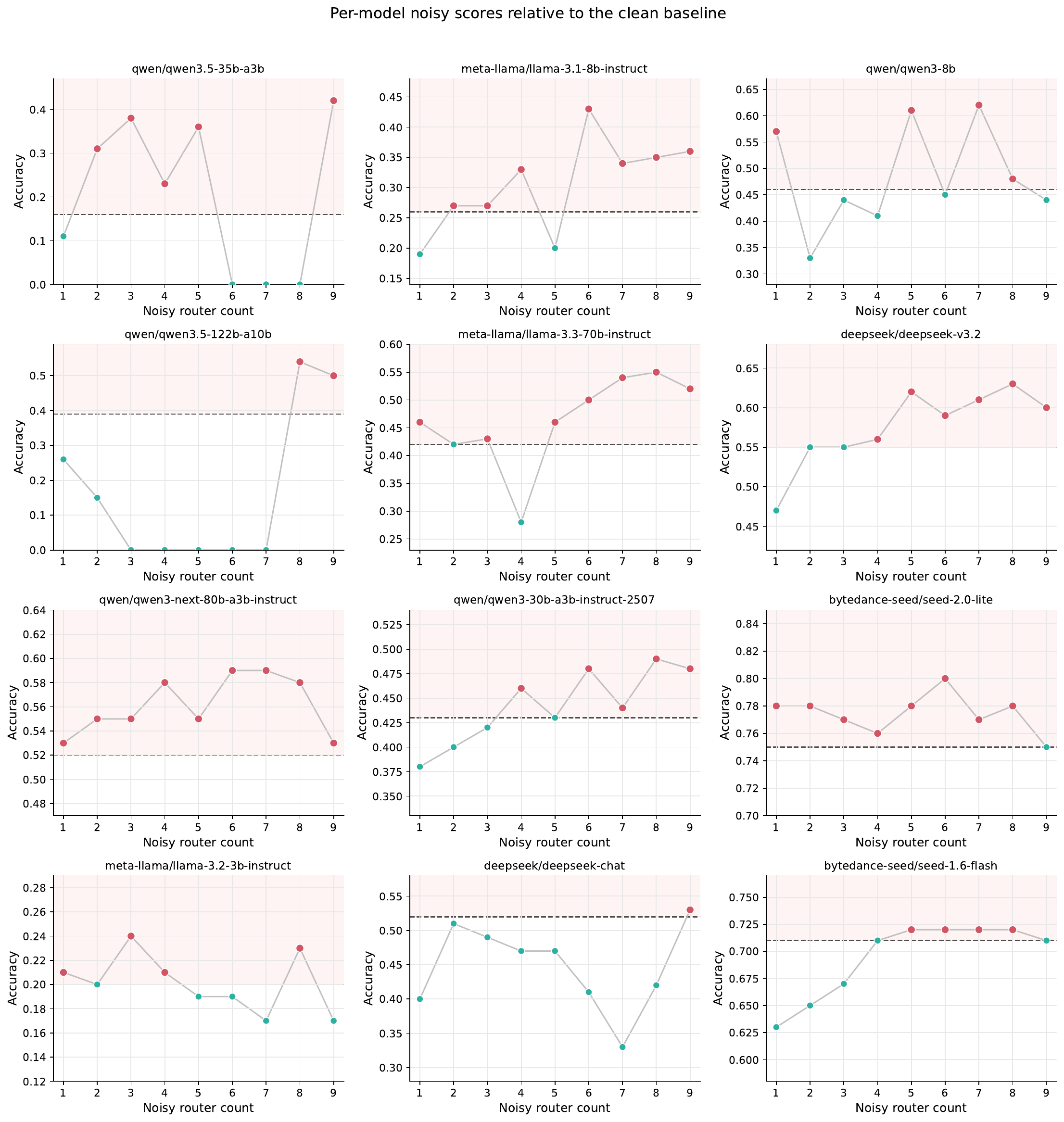}
  \caption{Per-model noisy scores relative to the clean baseline.}
  \label{fig:model_panels}
\end{figure*}
Figure~\ref{fig:model_panels} plots model-specific performance trajectories relative to the clean baseline under different noisy-router settings. The anomalous-gain pattern is clearly heterogeneous across models rather than uniformly distributed. Some models exceed the baseline under almost all noisy settings. For example, Qwen3-Next-80B violates the baseline in all 9 router settings, while Seed-2.0-Lite does so in 8 out of 9. By contrast, DeepSeek-Chat exceeds the baseline only once, and Qwen3.5-122B does so only twice.

The breadth and strength of anomalies are also not identical. Llama-3.1-8B and Llama-3.3-70B both violate the baseline in 7 out of 9 settings, indicating relatively broad exposure. By contrast, Qwen3.5-35B violates the baseline in only 5 settings but reaches a maximum positive excess of 0.260, the highest among all models. Qwen3-Next-80B displays a different pattern: violations occur in all 9 settings, but the maximum jump is smaller. This suggests that contamination sensitivity has at least two dimensions: how often a model crosses the baseline, and how far above the baseline it moves once it does so.

These results support Hypothesis~2. Even when models obtain similar benchmark scores, their sensitivity to contamination-related cues may differ substantially. Model comparison should therefore not stop at score levels alone, but should also consider whether those scores are equally credible. For space reasons, the compressed model-level summary is deferred to the appendix: Appendix Figure~\ref{fig:appendix_breadth} visualizes violation breadth across models, Appendix Table~\ref{tab:model_violation_summary} reports the corresponding model-level summary statistics, and Appendix Table~\ref{tab:model_router_details} provides the full model-by-router breakdown.

\subsection{Question-level mechanism: random fluctuation or directional improvement?}

\begin{figure*}[h]
  \centering
  \includegraphics[width=\textwidth]{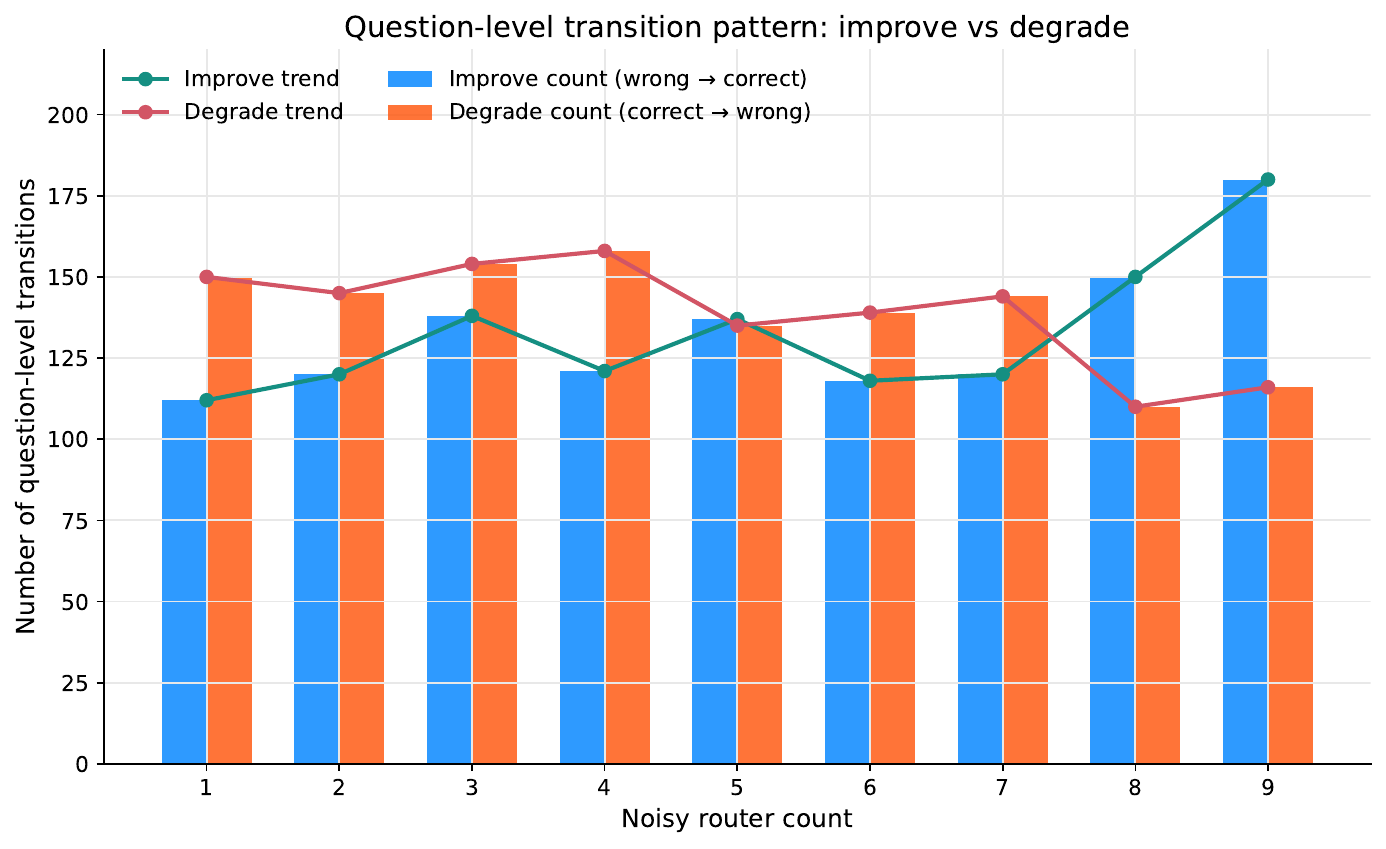}
  \caption{Question-level transition pattern: improve vs.\ degrade.}
  \label{fig:improve_degrade}
\end{figure*}
Figure~\ref{fig:improve_degrade} examines transition directions at the question level. Here, improve denotes a wrong\(\to\)correct transition, that is, a question answered incorrectly under the clean baseline but correctly under a noisy condition; degrade denotes a correct\(\to\)wrong transition. The figure aggregates these transitions across all models, so the vertical axis represents the total number of question-level transitions rather than the number of questions for any single model.The figure aggregates these transitions across all models, so the vertical axis represents the total number of question-level transitions rather than the number of questions for any single model.

The improve and degrade curves are not strictly monotonic at every router count, but their overall movement is clearly directional. As the number of noisy routers increases from 1 to 9, improve rises from 112 to 180, whereas degrade falls from 150 to 116. More importantly, their relative ordering reverses in higher-router regions. At \(r=1\), improve is 38 cases lower than degrade; by \(r=5\), the two are nearly balanced (137 versus 135); and by \(r=8\) and \(r=9\), improve reaches 150 and 180, clearly exceeding degrade at 110 and 116. In other words, as noisy routers accumulate, the question-level pattern shifts from ``more degraded than improved'' to ``more improved than degraded.''

This evidence is consistent with Hypothesis~3. The claim is not that every additional router mechanically raises improve and lowers degrade at each single point, but that the overall trend increasingly favors wrong\(\to\)correct transitions as router count grows, especially in higher-router conditions. This directional reversal suggests that noisy aggregation does not merely destroy information. It can also reconstruct semantically related cues that help recover benchmark answers, thereby turning previously incorrect responses into correct ones. The above-baseline anomaly is therefore reflected not only in score-level deviations but also in a question-level process of directional improvement.

\section{Conclusion and Discussion}

This paper develops an audit framework for interpreting benchmark scores through the lens of contamination sensitivity and score confidence in LLM evaluation. The empirical results reveal a coherent pattern across three levels. At the aggregate level, above-baseline anomalies under noisy conditions are not isolated exceptions but recur across multiple router settings. At the model level, the breadth of violations and the magnitude of positive excess differ substantially across models, indicating that benchmark scores vary in how sensitive they are to potential contamination-related cues. At the question level, as the number of noisy routers increases, wrong-to-correct transitions tend to rise, correct-to-wrong transitions tend to fall, and the former eventually exceeds the latter in higher-router conditions. Taken together, these findings suggest that benchmark scores are not merely straightforward records of answer correctness. They may also contain a nontrivial component related to how responsive a model is to benchmark-related semantic-neighbor information. A high score, therefore, does not necessarily imply a high-confidence score, and even models with similar benchmark performance may differ substantially in how credibly their scores represent genuine generalization ability.

The broader implication is conceptual as much as empirical. Current benchmark practice often compresses two qualitatively different capacities into a single numerical score. One is exam-oriented competence, that is, the ability to produce correct answers under fixed answer formats, input--output conventions, and judging rules. The other is principled capability, namely, the ability that remains after contamination and semantic leakage are excluded as much as possible. In practice, these two capacities are often treated as if they were the same thing. Our results suggest that this interpretation is too coarse. Benchmark performance should not automatically be read as a direct measure of genuine generalization, because part of what is being captured may reflect sensitivity to contamination-related cues rather than purely contamination-free capability. This is precisely why the problem of benchmark evaluation in the LLM era is not exhausted by asking which model scores higher; it must also include the question of what that score is actually measuring.

These findings do not imply that benchmarks are entirely invalid. For LLMs, the ability to retrieve and output the correct answer within an institutionalized answering structure is itself a real and practically relevant form of competence. In many applied settings, such competence has operational value, since real deployment environments often involve standardized prompts, constrained output formats, and repeated evaluation procedures. The problem, therefore, is not the existence of benchmark-based examinations themselves. Rather, the problem is the tendency to interpret institutional success within such examinations as direct evidence of pure generalization ability. In this sense, the core issue of Silicon Bureaucracy and AI Test-Oriented Education is not that models are being examined, but that examination outcomes are too easily treated as the whole of capability. Our argument is accordingly not to reject benchmarks, but to reinterpret them more carefully and to distinguish between score level and score credibility.

Recognizing this problem is not merely of theoretical interest; it is also useful and necessary in practice. Once contamination sensitivity is treated as an audit target, the framework proposed in this paper can be used as a practical diagnostic tool for benchmark governance. For model developers, especially frontier laboratories and commercial providers that continually update and retrain their models, a router--worker style audit can serve as an internal self-check during model iteration. Before releasing a new version, developers can test whether benchmark gains remain stable under deletion, rewriting, and perturbation, or whether part of the improvement is disproportionately driven by contamination-related semantic-neighbor cues. In this way, the framework can help distinguish genuine capability growth from score inflation that is overly dependent on benchmark familiarity.

The same logic is useful from the user side. For downstream users choosing among models with similar benchmark scores, raw score comparison alone may be insufficient. A model with a slightly lower score but weaker contamination sensitivity may in fact be more trustworthy than a model with a slightly higher score but stronger above-baseline anomalies under noisy conditions. The framework therefore provides a practical way to compare the credibility of seemingly similar scores and to support model selection in settings where reliability matters more than leaderboard optics. More generally, evaluation should move from a one-dimensional logic of ``who scores higher'' toward a two-dimensional logic that asks both ``how high is the score'' and ``how credible is the score as evidence of genuine capability.''

Public benchmarks have already become institutionalized examination and selection devices in the LLM ecosystem. Under such conditions, the meaning of a score can no longer be taken for granted. A benchmark score should not be treated as self-explanatory, because what it measures may be a mixture of exam-oriented competence and contamination-sensitive performance. Once this distinction is made explicit, benchmark evaluation no longer ends with the leaderboard. It must be supplemented by auditing procedures that clarify what part of the observed score is likely to reflect robust capability and what part may be entangled with benchmark-related semantic familiarity.

In sum, this paper argues that benchmark scores in the LLM era should be treated neither as meaningless nor as self-sufficient. They remain useful, but they are not transparent windows into genuine generalization. Once public benchmarks have become institutionalized examination and selection devices, the meaning of a score requires reinterpretation, and the use of a score requires additional auditing.

\bibliographystyle{ACM-Reference-Format}
\bibliography{main}

\appendix

\section{Technical appendices and supplementary material}

This appendix provides supplementary evidence that complements the main text at both the model level and the router level. It includes one additional figure and three additional tables. Together, these materials provide the full numerical background for the heterogeneity patterns discussed in Section~5.2 and the router-level and question-level transition results discussed in Sections~5.1 and~5.3.

Table~\ref{tab:model_violation_summary} reports model-level summary statistics, including violation breadth and positive-excess measures. 
{
\scriptsize
\begin{longtblr}[
  caption = {Model-level summary of violation breadth and positive excess},
  label   = {tab:model_violation_summary},
]{
  width   = \linewidth,      
  rowhead = 1,               
  colspec = {
    X[l]                     
    *{5}{S[table-format=2.2]}
  },
  row{1} = {font=\bfseries, guard}, 
  column{3-6} = {c},                
  colsep = 3pt,
  rowsep = 3pt,
}
\toprule
Model & {Viol.\\Count} & {Viol.\\Rate} & {Max Pos.\\Excess} & {Mean Pos.\\Excess} & {Mean\\Gain} \\
\midrule
Qwen3-Next-80B & 9/9 & 1.000 & 0.070 & 0.041 & 0.041 \\
Seed-2.0-Lite & 8/9 & 0.889 & 0.050 & 0.028 & 0.024 \\
Llama-3.1-8B & 7/9 & 0.778 & 0.170 & 0.076 & 0.044 \\
Llama-3.3-70B & 7/9 & 0.778 & 0.130 & 0.074 & 0.042 \\
DeepSeek-V3.2 & 6/9 & 0.667 & 0.080 & 0.052 & 0.026 \\
Qwen3.5-35B & 5/9 & 0.556 & 0.260 & 0.178 & 0.040 \\
Qwen3-30B & 5/9 & 0.556 & 0.060 & 0.040 & 0.012 \\
Qwen3-8B & 4/9 & 0.444 & 0.160 & 0.110 & 0.023 \\
Llama-3.2-3B & 4/9 & 0.444 & 0.040 & 0.022 & 0.001 \\
Seed-1.6-Flash & 4/9 & 0.444 & 0.010 & 0.010 & -0.016 \\
Qwen3.5-122B & 2/9 & 0.222 & 0.150 & 0.130 & -0.229 \\
DeepSeek-Chat & 1/9 & 0.111 & 0.010 & 0.010 & -0.072 \\
\bottomrule
\end{longtblr}
}






Table~\ref{tab:model_router_details} reports the full model-by-router breakdown underlying the anomalous-gain patterns discussed in the main text. 
{
\scriptsize
\begin{longtblr}[
  caption = {Model-by-router violation details},
  label   = {tab:model_router_details},
]{
  width   = \linewidth,      
  rowhead = 1,               
  colspec = {
    X[l]                     
    *{6}{S[table-format=2.2]}
  },
  row{1} = {font=\bfseries, guard}, 
  column{3-6} = {c},                
  colsep = 3pt,
  rowsep = 3pt,
}
\toprule
Model & Router & Clean & Noisy & Gain & Improve & Degrade \\
\midrule
Seed-1.6-Flash & 1 & 0.71 & 0.63 & -0.08 & 8 & 16 \\
Seed-1.6-Flash & 2 & 0.71 & 0.65 & -0.06 & 6 & 12 \\
Seed-1.6-Flash & 3 & 0.71 & 0.67 & -0.04 & 8 & 12 \\
Seed-1.6-Flash & 4 & 0.71 & 0.71 & 0.00 & 6 & 6 \\
Seed-1.6-Flash & 5 & 0.71 & 0.72 & 0.01 & 7 & 6 \\
Seed-1.6-Flash & 6 & 0.71 & 0.72 & 0.01 & 6 & 5 \\
Seed-1.6-Flash & 7 & 0.71 & 0.72 & 0.01 & 6 & 5 \\
Seed-1.6-Flash & 8 & 0.71 & 0.72 & 0.01 & 5 & 4 \\
Seed-1.6-Flash & 9 & 0.71 & 0.71 & 0.00 & 6 & 6 \\
Seed-2.0-Lite & 1 & 0.75 & 0.78 & 0.03 & 6 & 3 \\
Seed-2.0-Lite & 2 & 0.75 & 0.78 & 0.03 & 6 & 3 \\
Seed-2.0-Lite & 3 & 0.75 & 0.77 & 0.02 & 5 & 3 \\
Seed-2.0-Lite & 4 & 0.75 & 0.76 & 0.01 & 5 & 4 \\
Seed-2.0-Lite & 5 & 0.75 & 0.78 & 0.03 & 4 & 1 \\
Seed-2.0-Lite & 6 & 0.75 & 0.80 & 0.05 & 5 & 0 \\
Seed-2.0-Lite & 7 & 0.75 & 0.77 & 0.02 & 4 & 2 \\
Seed-2.0-Lite & 8 & 0.75 & 0.78 & 0.03 & 5 & 2 \\
Seed-2.0-Lite & 9 & 0.75 & 0.75 & 0.00 & 5 & 5 \\
DeepSeek-Chat & 1 & 0.52 & 0.40 & -0.12 & 5 & 17 \\
DeepSeek-Chat & 2 & 0.52 & 0.51 & -0.01 & 10 & 11 \\
DeepSeek-Chat & 3 & 0.52 & 0.49 & -0.03 & 13 & 16 \\
DeepSeek-Chat & 4 & 0.52 & 0.47 & -0.05 & 10 & 15 \\
DeepSeek-Chat & 5 & 0.52 & 0.47 & -0.05 & 9 & 14 \\
DeepSeek-Chat & 6 & 0.52 & 0.41 & -0.11 & 9 & 20 \\
DeepSeek-Chat & 7 & 0.52 & 0.33 & -0.19 & 6 & 25 \\
DeepSeek-Chat & 8 & 0.52 & 0.42 & -0.10 & 7 & 17 \\
DeepSeek-Chat & 9 & 0.52 & 0.53 & 0.01 & 11 & 10 \\
DeepSeek-V3.2 & 1 & 0.55 & 0.47 & -0.08 & 7 & 15 \\
DeepSeek-V3.2 & 2 & 0.55 & 0.55 & 0.00 & 9 & 9 \\
DeepSeek-V3.2 & 3 & 0.55 & 0.55 & 0.00 & 11 & 11 \\
DeepSeek-V3.2 & 4 & 0.55 & 0.56 & 0.01 & 10 & 9 \\
DeepSeek-V3.2 & 5 & 0.55 & 0.62 & 0.07 & 17 & 10 \\
DeepSeek-V3.2 & 6 & 0.55 & 0.59 & 0.04 & 14 & 10 \\
DeepSeek-V3.2 & 7 & 0.55 & 0.61 & 0.06 & 13 & 7 \\
DeepSeek-V3.2 & 8 & 0.55 & 0.63 & 0.08 & 16 & 8 \\
DeepSeek-V3.2 & 9 & 0.55 & 0.60 & 0.05 & 13 & 8 \\
Llama-3.1-8B & 1 & 0.26 & 0.19 & -0.07 & 6 & 13 \\
Llama-3.1-8B & 2 & 0.26 & 0.27 & 0.01 & 11 & 10 \\
Llama-3.1-8B & 3 & 0.26 & 0.27 & 0.01 & 17 & 16 \\
Llama-3.1-8B & 4 & 0.26 & 0.33 & 0.07 & 17 & 10 \\
Llama-3.1-8B & 5 & 0.26 & 0.20 & -0.06 & 11 & 17 \\
Llama-3.1-8B & 6 & 0.26 & 0.43 & 0.17 & 21 & 4 \\
Llama-3.1-8B & 7 & 0.26 & 0.34 & 0.08 & 18 & 10 \\
Llama-3.1-8B & 8 & 0.26 & 0.35 & 0.09 & 18 & 9 \\
Llama-3.1-8B & 9 & 0.26 & 0.36 & 0.10 & 17 & 7 \\
Llama-3.2-3B & 1 & 0.20 & 0.21 & 0.01 & 7 & 6 \\
Llama-3.2-3B & 2 & 0.20 & 0.20 & 0.00 & 5 & 5 \\
Llama-3.2-3B & 3 & 0.20 & 0.24 & 0.04 & 10 & 6 \\
Llama-3.2-3B & 4 & 0.20 & 0.21 & 0.01 & 9 & 8 \\
Llama-3.2-3B & 5 & 0.20 & 0.19 & -0.01 & 4 & 5 \\
Llama-3.2-3B & 6 & 0.20 & 0.19 & -0.01 & 7 & 8 \\
Llama-3.2-3B & 7 & 0.20 & 0.17 & -0.03 & 9 & 12 \\
Llama-3.2-3B & 8 & 0.20 & 0.23 & 0.03 & 8 & 5 \\
Llama-3.2-3B & 9 & 0.20 & 0.17 & -0.03 & 8 & 11 \\
Llama-3.3-70B & 1 & 0.42 & 0.46 & 0.04 & 10 & 6 \\
Llama-3.3-70B & 2 & 0.42 & 0.42 & 0.00 & 11 & 11 \\
Llama-3.3-70B & 3 & 0.42 & 0.43 & 0.01 & 15 & 14 \\
Llama-3.3-70B & 4 & 0.42 & 0.28 & -0.14 & 5 & 19 \\
Llama-3.3-70B & 5 & 0.42 & 0.46 & 0.04 & 15 & 11 \\
Llama-3.3-70B & 6 & 0.42 & 0.50 & 0.08 & 15 & 7 \\
Llama-3.3-70B & 7 & 0.42 & 0.54 & 0.12 & 17 & 5 \\
Llama-3.3-70B & 8 & 0.42 & 0.55 & 0.13 & 17 & 4 \\
Llama-3.3-70B & 9 & 0.42 & 0.52 & 0.10 & 17 & 7 \\
Qwen3-30B & 1 & 0.43 & 0.38 & -0.05 & 7 & 12 \\
Qwen3-30B & 2 & 0.43 & 0.40 & -0.03 & 9 & 12 \\
Qwen3-30B & 3 & 0.43 & 0.42 & -0.01 & 9 & 10 \\
Qwen3-30B & 4 & 0.43 & 0.46 & 0.03 & 13 & 10 \\
Qwen3-30B & 5 & 0.43 & 0.43 & 0.00 & 7 & 7 \\
Qwen3-30B & 6 & 0.43 & 0.48 & 0.05 & 13 & 8 \\
Qwen3-30B & 7 & 0.43 & 0.44 & 0.01 & 10 & 9 \\
Qwen3-30B & 8 & 0.43 & 0.49 & 0.06 & 14 & 8 \\
Qwen3-30B & 9 & 0.43 & 0.48 & 0.05 & 15 & 10 \\
Qwen3-8B & 1 & 0.46 & 0.57 & 0.11 & 23 & 12 \\
Qwen3-8B & 2 & 0.46 & 0.33 & -0.13 & 12 & 25 \\
Qwen3-8B & 3 & 0.46 & 0.44 & -0.02 & 15 & 17 \\
Qwen3-8B & 4 & 0.46 & 0.41 & -0.05 & 17 & 22 \\
Qwen3-8B & 5 & 0.46 & 0.61 & 0.15 & 22 & 7 \\
Qwen3-8B & 6 & 0.46 & 0.45 & -0.01 & 16 & 17 \\
Qwen3-8B & 7 & 0.46 & 0.62 & 0.16 & 24 & 8 \\
Qwen3-8B & 8 & 0.46 & 0.48 & 0.02 & 20 & 18 \\
Qwen3-8B & 9 & 0.46 & 0.44 & -0.02 & 17 & 19 \\
Qwen3-Next-80B & 1 & 0.52 & 0.53 & 0.01 & 8 & 7 \\
Qwen3-Next-80B & 2 & 0.52 & 0.55 & 0.03 & 8 & 5 \\
Qwen3-Next-80B & 3 & 0.52 & 0.55 & 0.03 & 7 & 4 \\
Qwen3-Next-80B & 4 & 0.52 & 0.58 & 0.06 & 11 & 5 \\
Qwen3-Next-80B & 5 & 0.52 & 0.55 & 0.03 & 11 & 8 \\
Qwen3-Next-80B & 6 & 0.52 & 0.59 & 0.07 & 12 & 5 \\
Qwen3-Next-80B & 7 & 0.52 & 0.59 & 0.07 & 13 & 6 \\
Qwen3-Next-80B & 8 & 0.52 & 0.58 & 0.06 & 11 & 5 \\
Qwen3-Next-80B & 9 & 0.52 & 0.53 & 0.01 & 7 & 6 \\
Qwen3.5-122B & 1 & 0.39 & 0.26 & -0.13 & 15 & 28 \\
Qwen3.5-122B & 2 & 0.39 & 0.15 & -0.24 & 8 & 32 \\
Qwen3.5-122B & 3 & 0.39 & 0.00 & -0.39 & 0 & 39 \\
Qwen3.5-122B & 4 & 0.39 & 0.00 & -0.39 & 0 & 39 \\
Qwen3.5-122B & 5 & 0.39 & 0.00 & -0.39 & 0 & 39 \\
Qwen3.5-122B & 6 & 0.39 & 0.00 & -0.39 & 0 & 39 \\
Qwen3.5-122B & 7 & 0.39 & 0.00 & -0.39 & 0 & 39 \\
Qwen3.5-122B & 8 & 0.39 & 0.54 & 0.15 & 29 & 14 \\
Qwen3.5-122B & 9 & 0.39 & 0.50 & 0.11 & 26 & 15 \\
Qwen3.5-35B & 1 & 0.16 & 0.11 & -0.05 & 10 & 15 \\
Qwen3.5-35B & 2 & 0.16 & 0.31 & 0.15 & 25 & 10 \\
Qwen3.5-35B & 3 & 0.16 & 0.38 & 0.22 & 28 & 6 \\
Qwen3.5-35B & 4 & 0.16 & 0.23 & 0.07 & 18 & 11 \\
Qwen3.5-35B & 5 & 0.16 & 0.36 & 0.20 & 30 & 10 \\
Qwen3.5-35B & 6 & 0.16 & 0.00 & -0.16 & 0 & 16 \\
Qwen3.5-35B & 7 & 0.16 & 0.00 & -0.16 & 0 & 16 \\
Qwen3.5-35B & 8 & 0.16 & 0.00 & -0.16 & 0 & 16 \\
Qwen3.5-35B & 9 & 0.16 & 0.42 & 0.26 & 38 & 12 \\
\bottomrule
\end{longtblr}
}

Table~\ref{tab:router_level_summary} reports router-level violation statistics together with the corresponding question-level transition counts.

{
\scriptsize
\begin{longtblr}[
  caption = {Router-level violation statistics and transition counts},
  label   = {tab:router_level_summary},
]{
  width   = \linewidth,      
  rowhead = 1,               
  colspec = {
    *{7}{S[table-format=2.2]}
  },
  row{1} = {font=\bfseries, guard}, 
  column{3-6} = {c},                
  colsep = 3pt,
  rowsep = 3pt,
}
\toprule
Router & {Viol.\\Models} & {Viol.\\Rate} & {Mean Pos.\\Excess} & {Improve} & {Degrade} & {Net\\Improve} \\
\midrule

1 & 5/12 & 0.417 & 0.040 & 112 & 150 & -38 \\
2 & 4/12 & 0.333 & 0.055 & 120 & 145 & -25 \\
3 & 6/12 & 0.500 & 0.055 & 138 & 154 & -16 \\
4 & 7/12 & 0.583 & 0.037 & 121 & 158 & -37 \\
5 & 7/12 & 0.583 & 0.076 & 137 & 135 & 2 \\
6 & 7/12 & 0.583 & 0.067 & 118 & 139 & -21 \\
7 & 8/12 & 0.667 & 0.066 & 120 & 144 & -24 \\
8 & 10/12 & 0.833 & 0.066 & 150 & 110 & 40 \\
9 & 8/12 & 0.667 & 0.086 & 180 & 116 & 64 \\

\bottomrule
\end{longtblr}
}





\begin{figure}[h]
\centering
\includegraphics[width=0.5\textwidth]{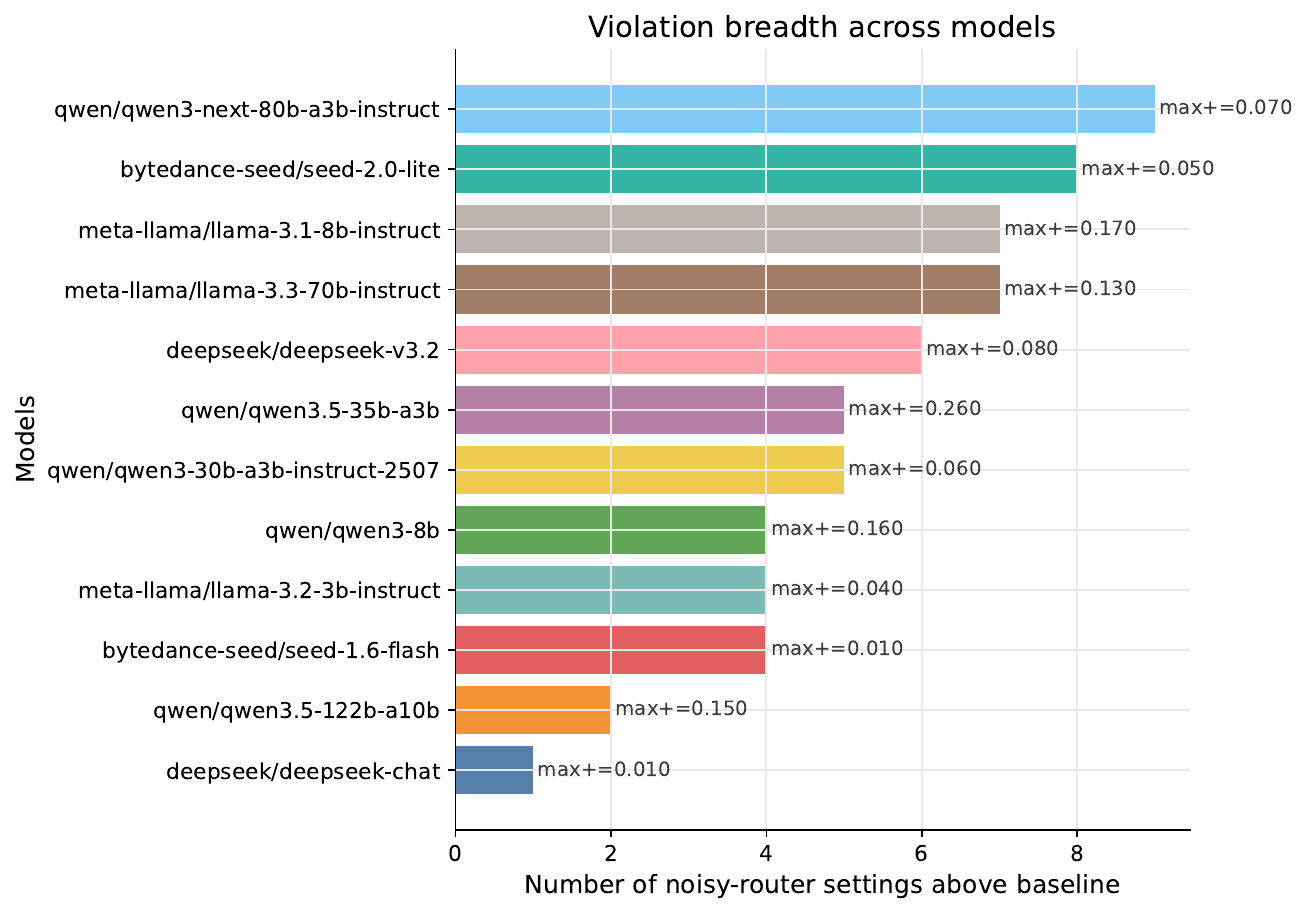}
\caption{Violation breadth across models.}
\label{fig:appendix_breadth}
\end{figure}

Figure~\ref{fig:appendix_breadth} provides a compact model-level visualization of violation breadth. 

\end{document}